\pdfoutput=1

\documentclass[11pt]{article}

\usepackage{acl}

\usepackage{times}
\usepackage{latexsym}

\usepackage[T1]{fontenc}

\usepackage[utf8]{inputenc}

\usepackage{microtype}

\usepackage{multirow}
\usepackage{amsmath}
\usepackage{cleveref}
\usepackage{booktabs}
\usepackage{mathrsfs}
\usepackage{graphicx}
\usepackage{bbding}
\usepackage[shortlabels]{enumitem}
\usepackage[noorphans,vskip=0.5ex,leftmargin=1em]{quoting}
\setlist{nosep}

\title{Transformer-Based Temporal Information Extraction and Application: A Review}

\author{Xin Su$^{1}$  \quad 
        Phillip Howard$^{1}$ \quad 
        Steven Bethard$^{2}$ \\
        $^{1}$Intel Labs \quad $^{2}$University of Arizona \\
        {\tt {\{xin.su, phillip.r.howard\}}@arizona.edu}, {\tt bethard@arizona.edu}} 

\begin{document}
\maketitle

\begin{abstract}
Temporal information extraction (IE) aims to extract structured temporal information from unstructured text, thereby uncovering the implicit timelines within. This technique is applied across domains such as healthcare, newswire, and intelligence analysis, aiding models in these areas to perform temporal reasoning and enabling human users to grasp the temporal structure of text. Transformer-based pre-trained language models have produced revolutionary advancements in natural language processing, demonstrating exceptional performance across a multitude of tasks. Despite the achievements garnered by Transformer-based approaches in temporal IE, there is a lack of comprehensive reviews on these endeavors. In this paper, we aim to bridge this gap by systematically summarizing and analyzing the body of work on temporal IE using Transformers while highlighting potential future research directions.
\end{abstract}

\section{Introduction}
\label{sec:introduction}
\begin{figure}
    \centering
    \includegraphics[width=\columnwidth]{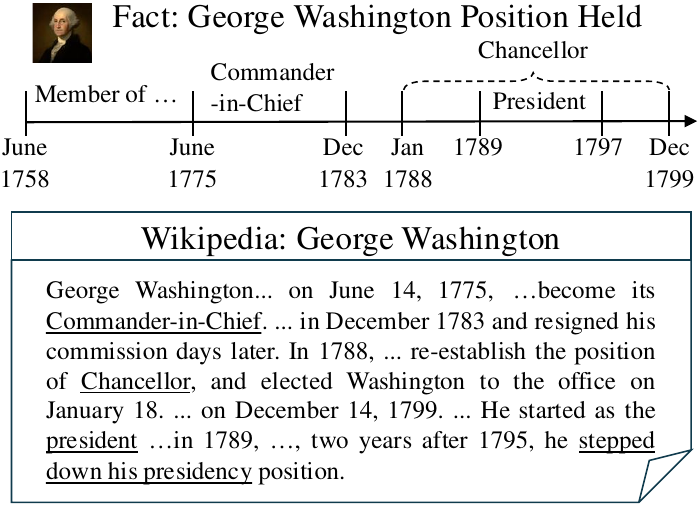}
    \caption{A snippet from George Washington's Wikipedia page and a timeline regarding his positions.}
    \label{fig:timeline-example}
\end{figure}

Temporal information extraction (IE) is a critical task in natural language processing (NLP). Its objective is to extract structured temporal information from unstructured text, thereby revealing the implicit timelines within the text. This not only helps improve temporal reasoning in other NLP tasks, such as timeline summarization and temporal question answering, but also helps human users in gaining a deeper understanding of the evolution of text content over time. For example, \Cref{fig:timeline-example} displays a snippet of George Washington's Wikipedia page and the timeline of his position changes; relying solely on text-heavy documents to trace his position changes over different years is time-consuming and may lack accuracy as facts and temporal expressions are scattered throughout the text. 
In contrast, a timeline enables both NLP models and humans to understand the changes in these positions over time more succinctly and clearly. The application of this structured temporal information is not limited to Wikipedia but is also widely used in other domains such as healthcare \cite{styler-iv-etal-2014-temporal}.

The advent of the Transformer architecture \cite{vaswani2017attention} has sparked a revolutionary change in the field of NLP, particularly with the recent Transformer-based generative large language models (LLM), such as LLAMA3 \cite{dubey2024llama} and GPT-4 \cite{achiam2023gpt}, demonstrating exceptional performance across many tasks. Nevertheless, there has yet to be an in-depth study that provides a comprehensive review or analysis of the Transformer architecture's application in the field of temporal IE. Existing surveys \cite{lim2019survey,leeuwenberg2019survey,alfattni2020extraction,olex2021review} focus on rule-based systems or traditional machine learning models (e.g., support vector machines) which are reliant on hand-crafted features. Only \citet{olex2021review} touches on the application of Transformer models, but they offer only a brief description of BERT-style models and focus largely on the clinical domain.

To address this gap, we systematically review the applications of Transformer-based models in the field of temporal IE. Broadly, temporal IE refers to any tasks involving the extraction of temporal information from text. We focus on three important tasks which are defined in the most widely adopted temporal IE annotation framework, TimeML \cite{james2003timeml}: time expression identification,  time expression normalization, and temporal relation extraction.  Our contributions are summarized as follows:
(1) We systematically review, summarize, and categorize the existing temporal IE datasets, Transformer-based methods, and applications.
(2) We identify and highlight the research gaps in the field of temporal IE and suggest potential directions for future research.

\section{Overview}
\label{sec:preliminary}
\begin{figure}
    \centering
    \includegraphics[width=\columnwidth]{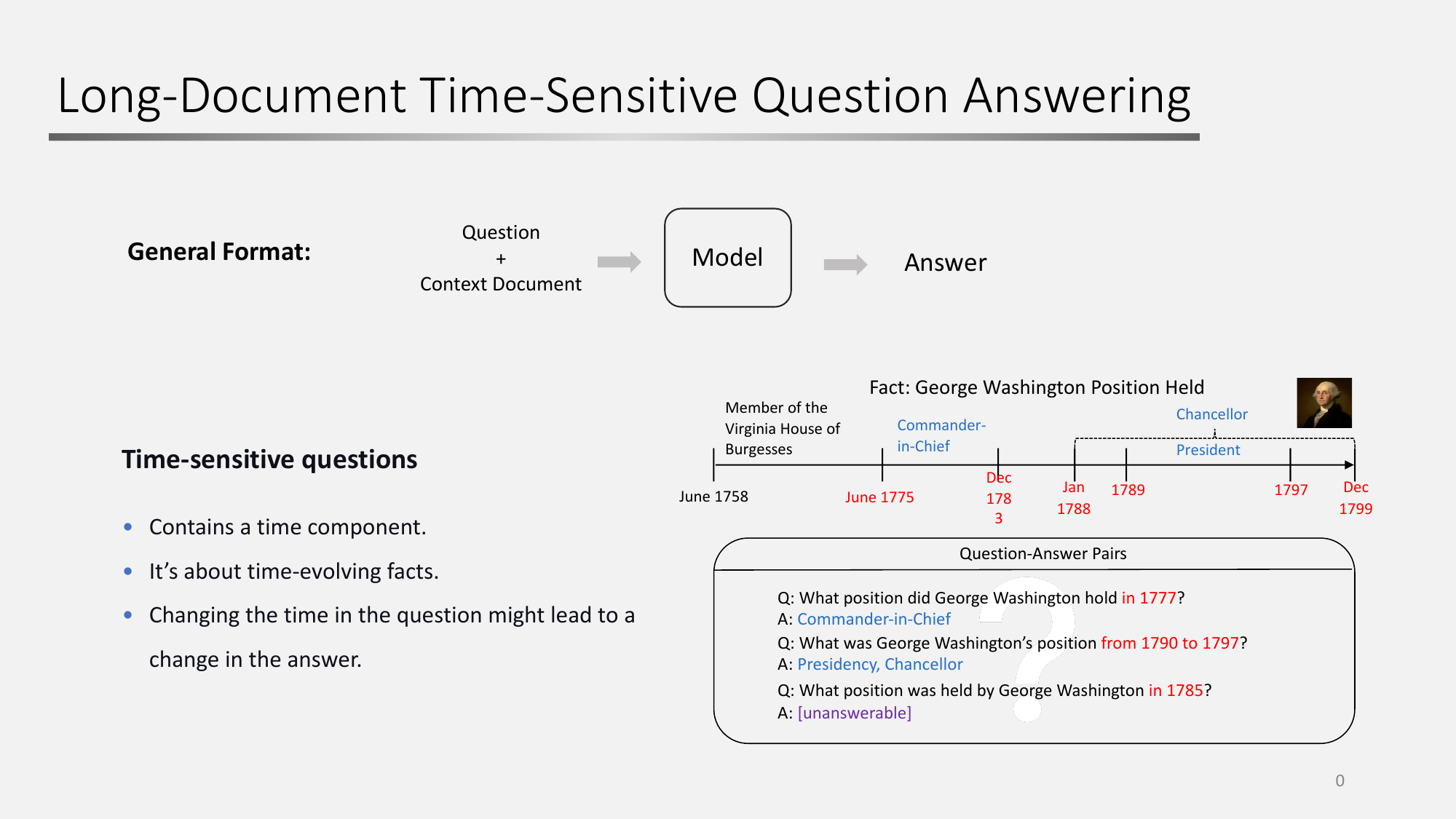}
    \caption{A snippet from George Washington's Wikipedia page and the corresponding temporal graph.}
    \label{fig:temporal-graph}
\end{figure}

The goal of temporal IE is to extract structured temporal information from unstructured text, facilitating its interpretation and processing by computers, thereby achieving a transformation from text to structure. The final result of a temporal IE system is the construction of a directed acyclic graph, or a temporal graph, which represents the structured temporal information in the text. In the temporal graph, nodes represent time expressions and events
(temporal entities), while edges depict the temporal relations between these nodes, such as ``before,'' ``after,'' etc. For instance, \Cref{fig:temporal-graph} illustrates a text snippet from George Washington's Wikipedia page and its corresponding temporal graph.

Constructing a temporal graph involves several sub-tasks: time expression identification, time expression normalization, event extraction, and temporal relation extraction. The following is a brief introduction to these sub-tasks; see \Cref{sec:eval-metrics} for a discussion of common evaluation methods.


\paragraph{Time Expression Identification and Normalization}
Time expression identification refers to identifying specific time points, durations, or periods within the text, such as the explicitly dateable expression ``February 25, 2024,'' or more ambiguous expressions like ``three days ago'' \cite{james2003timeml}. Time normalization involves converting identified expressions into a standardized format to improve their interpretability. For example, under the ISO-TimeML framework \cite{pustejovsky-etal-2010-iso}, ``February 25, 2024'' might be converted into the TIMEX3 format as ``2024-02-25''.

\paragraph{Event Trigger Extraction}
In temporal IE, event extraction differs from other NLP event extraction tasks; it simply marks the event trigger words that represent actions, such as ``accident'' in ``about two weeks after the accident occurred''. We will not review event extraction works because, to our knowledge, there is currently no temporal IE research focused solely on event extraction. Furthermore, most existing work on temporal IE assumes that event triggers have already been identified. For a comprehensive survey of event extraction, we refer readers to \cite{li2022survey}.

\paragraph{Temporal Relation Extraction}
The task of temporal relation extraction aims to identify the temporal relations among given events and time expressions. Common temporal relations include before, after, and simultaneous. For example, in \Cref{fig:temporal-graph}, the temporal relation between ``June 14, 1775'' and the event ``become'' is marked as ``after''.

\section{Datasets}
\label{sec:temporal-datasets}

A clearly defined annotation framework is essential when constructing a dataset for temporal IE. It needs to precisely define time expressions, events, and their relations. We summarize all the datasets in \Cref{table:dataset-overview} of \Cref{sec:datasets-summary}.

\subsection{TimeML Annotation Framework Datasets}
An end-to-end temporal IE dataset encompasses various tasks, including the identification and normalization of time expressions and the extraction of temporal relations. Most end-to-end temporal information datasets have been based on the TimeML framework \cite{james2003timeml} or its derivatives, such as ISO-TimeML \cite{pustejovsky-etal-2010-iso}. We present datasets based on the TimeML framework in the first section of \Cref{table:dataset-overview}.

TimeBank \cite{james2003timeml} was the first dataset to adopt the TimeML framework, focusing on the English news domain. Follow-up works included the TempEval shared task series \cite{verhagen-etal-2007-semeval,verhagen-etal-2010-semeval,uzzaman-etal-2013-semeval}, covering multiple languages, including Chinese, English, Italian, French, Korean, and Spanish. There are also language-specific datasets like French TimeBank \cite{bittar-etal-2011-french}, Spanish TimeBank \cite{nieto2011modes}, Portuguese TimeBank \cite{costa-branco-2012-timebankpt}, Japanese TimeBank \cite{asahara-etal-2013-bccwj}, Italian TimeBank \cite{bracchi2016enrichring}, and Korean TimeBank \cite{lim-etal-2018-korean}. Similarly, the MeanTime dataset \cite{minard-etal-2016-meantime} offers data in English, Italian, Spanish, and Dutch. Datasets based on TimeML and its variants showcase language diversity and also cover several different domains: the Spanish TimeBank focuses on history text, the Korean TimeBank is based on Wikipedia content, and the Richer Event Description dataset \cite{ogorman-etal-2016-richer} provides data from both news and forum discussion domains.

Additionally, efforts have been made to improve the temporal relation annotations in the original TimeBank. TimeBank-Dense \cite{chambers-etal-2014-dense} addresses the sparsity of temporal relation annotations in TimeBank by requiring annotators to label all temporal relations within a given scope, thus increasing the number of temporal relations in the dataset. The TORDER dataset \cite{cheng-miyao-2018-inducing} annotates the same documents as TimeBank-Dense, introducing temporal relations automatically by anchoring times and events to absolute points, reducing the annotation burden. The MATRES dataset \cite{ning-etal-2018-multi} focuses on events from TimeBank-Dense, anchoring events to different timelines and comparing their start times to enhance inter-annotator consistency.

Several datasets have been developed specific to the clinical domain, of which the Thyme datasets \cite{bethard-etal-2015-semeval,bethard-etal-2016-semeval,bethard-etal-2017-semeval} are most notable. They are based on the Thyme-TimeML \cite{styler-iv-etal-2014-temporal} annotation framework, which adjusts and adds new temporal attributes from ISO-TimeML to suit medical texts. Like the TimeBank series, the Thyme dataset involves identifying and normalizing time expressions and extracting temporal relations, focusing on English. Another similar dataset is i2b2-2012 \cite{sun2013evaluating}, which adapts the TimeML framework for clinical texts.

Besides end-to-end datasets, several others based on TimeML or its variants focus on specific temporal IE tasks. For instance, the AncientTimes dataset \cite{strotgen-etal-2014-extending} covers a broad range of languages, concentrating on the identification and normalization of time expressions. The TDDiscourse dataset \cite{naik-etal-2019-tddiscourse}, based on TimeBank-Dense, expands the annotation window for temporal relations, focusing on their extraction. The German time expression \cite{strotgen-etal-2018-krauts} and German VTEs \cite{may-etal-2021-extraction} datasets are dedicated to identifying and normalizing time expressions in German. The PATE dataset \cite{zarcone-etal-2020-pate} provides data aimed at time expression identification and normalization for the virtual assistant domain.

\subsection{Other Annotation Framework Datasets}
Unlike datasets for temporal IE based on TimeML, other annotation frameworks typically focus on specific sub-tasks of temporal IE, such as time expression identification and normalization or the extraction of temporal relations. We present these datasets in the second section of \Cref{table:dataset-overview}.

For time expression identification and normalization, WikiWars \cite{mazur-dale-2010-wikiwars} and SCATE \cite{laparra-etal-2018-semeval} are two major datasets. WikiWars contains data from English and German Wikipedia, annotated based on TIMEX2 (a precursor to TimeML's TIMEX3) to mark explicit time expressions. The SCATE dataset, based on English news and clinical documents, aims to address limitations in TimeML that prevent expressing multiple calendar units, times relative to events, and compositional time expressions. To achieve this, SCATE represents time expressions as compositions of temporal operators.

For temporal relations, there are datasets based on the temporal dependency tree/graph \cite{zhang-xue-2018-structured,zhang-xue-2019-acquiring,yao-etal-2020-annotating} and CaTeRS \cite{mostafazadeh-etal-2016-caters} frameworks. Unlike the pairwise temporal relations considered in the TimeML framework, temporal dependency tree assumes that all time expressions and events in a document have a reference time, allowing for the representation of overall temporal relations through a dependency tree. The subsequent temporal dependency graph dataset \cite{yao-etal-2020-annotating} relaxed this assumption by enabling each event in a document to have a reference event, a reference time, or both, thus forming a temporal graph structure. The temporal dependency tree dataset covers news and narrative domains in English and Chinese, while the temporal dependency graph dataset focuses on English news. Meanwhile, CaTeRS concentrates on analyzing temporal relations between events in English commonsense stories, with event definitions based on ontologies, different from the verb-, adjective-, or noun-based definitions in TimeML. CaTeRS' annotation of temporal relations is story-wide, with a simplified set of relations.


\subsection{Discussion and Research Gaps}
\label{sec:datasets-gaps}
\paragraph{Domain Bias} Existing annotated datasets exhibit significant domain biases. As demonstrated in \Cref{table:dataset-overview}, among the 32 datasets we reviewed, 20 (or 63\%) are predominantly focused on the newswire domain. While temporal information is crucial for understanding news content, an excessive concentration in a single domain hampers the advancement and generalizability of systems trained on these datasets, since the challenges and difficulties encountered in temporal IE vary across different domains. 
Notably, the Clinical TempEval 2017 shared task \citep{bethard-etal-2017-semeval} reveals that most tasks suffer an approximately 20-point drop in performance in a cross-domain setting, underscoring how domain shifts can significantly degrade model accuracy.
For example, temporal information, especially time expressions, in newswire texts tend to be explicitly stated, whereas in other domains, like historical Wikipedia entries, they might appear in subtler ways. Consider a statement from a page about George Washington that reads, ``\ldots 1798, one year after that, he stepped down from the presidency,'' which would demand a more nuanced interpretation for accurate time normalization. 
Cultivating datasets that represent a variety of domains is vital to driving innovation in temporal IE.

\paragraph{Language Diversity} 
Unlike the domain homogeneity of the datasets, the existing datasets display rich linguistic diversity, covering 15 different languages. The representation of time varies across languages, and even when semantically similar, the specific time intervals on the timeline can differ. For example, analysis in \citet{shwartz-2022-good} shows that different cultures/languages have significant variations in the understanding of ``night'' and ``evening'' during the day. One instance is that Brazilian Portuguese speakers often use ``evening'' and ``night'' interchangeably to denote the same time period, possibly because the tropical climate in Brazil causes evening to transition quickly into night. However, this might not be applicable to other cultures or languages. Therefore, the language diversity in datasets is crucial for developing models capable of effectively extracting temporal information across different languages.

\paragraph{Annotation and Dataset Framework Development Slows Down}
Aside from the original TimeML and some incremental modifications to it, no new end-to-end temporal IE annotation frameworks have been proposed. A significant issue with the existing TimeML-based annotation frameworks is the limited amount of information that the resultant temporal graphs can represent. For instance, in \Cref{fig:temporal-graph}, we only see trigger words for events, time expressions, and some temporal relations. When these temporal graphs are isolated from their original context and treated as stand-alone entities, they struggle to provide a comprehensive understanding of the textual information. This might explain why, in the upcoming \Cref{sec:Temporal-Information-Extraction-Application}, we see no work directly employing these extracted temporal graphs for reasoning to accomplish specific tasks, such as answering temporal questions. Instead, these temporal graphs are used as auxiliary tools or additional knowledge to assist task-specific models in temporal reasoning. 

In addition to the stagnation in the innovation of end-to-end annotation frameworks, there has been a notable decline in dataset development efforts in the field of temporal IE in recent years. This trend may primarily stem from the intrinsic complexity of the annotation process for temporal IE datasets. Such complexity accounts for the low annotator agreement observed in many annotation tasks \cite{cassidy-etal-2014-annotation}. Furthermore, as demonstrated by analysis in \citet{su-etal-2021-university}, even Ph.D. students in relevant fields find it challenging to comprehend annotation guidelines and annotate high-quality data within a short period. These issues highlight the difficulties in developing temporal IE datasets, suggesting that improvements in the annotation framework might be necessary to address these challenges.

\section{Time Expression Methods}
\label{sec:time-expression}
\subsection{Methods Overview}
In the realm of time expression identification, most prior work \cite{almasian2021bert,chen2019exploring,mirzababaei-etal-2022-hengam,olex2022temporal,laparra-etal-2021-semeval,almasian2022time,cao-etal-2022-xltime} leverages discriminative models built upon transformer encoders like BERT \cite{devlin-etal-2019-bert}. These approaches typically frame time expression identification as a token classification task, wherein a sequence of tokens is input, processed through a base encoder model to obtain contextualized representations, and  these representations are fed into a classifier (such as a simple linear classification layer or a Conditional Random Field layer) to identify time expressions and their specific types. \citet{almasian2021bert} is the only work exploring a generative approach for time expression identification, framing the task as a sequence-to-sequence problem and employing a pair of transformer encoders to formulate an encoder-decoder model—where one serves as the encoder and the other as the decoder—to generate additional TIMEX3 tags for the input, thereby recognizing time expressions and their types.

\citet{shwartz-2022-good} and \citet{kim-etal-2020-time} focus on the normalization of time expressions and use transformer-based models. \citet{shwartz-2022-good} aims to normalize time expressions from various cultural contexts (e.g., morning, noon, afternoon) into precise hourly representations within a day. They train a BERT model with a masked language modeling task to predict specific times of day that are masked, given the time expressions. \citet{kim-etal-2020-time} seeks to normalize time expressions in novels into specific daily hours, fine-tuning the BERT model for a 24-class classification task to ascertain the corresponding times of day for given expressions.

\citet{lange-etal-2023-multilingual} addresses both extraction and normalization of time expressions, adopting a pipeline approach. Initially, they fine-tune the XLM-R model using the token classification method to extract time expressions, then denote identified expressions with TIMEX3 tags with masked time values, and finally fine-tune the XLM-R model with masked language modeling to predict the normalized masked time values.

Several of the aforementioned works also utilize data augmentation techniques to improve the model's multilingual performance \cite{lange-etal-2023-multilingual,mirzababaei-etal-2022-hengam,almasian2022time}. For instance, \citet{lange-etal-2023-multilingual} employs the rule-based HeidelTime method \cite{strotgen-gertz-2010-heideltime} to annotate time expressions and their normalizations across 87 languages, generating a semi-supervised dataset to facilitate model training.

\subsection{Discussion and Research Gaps}
\label{sec:time-expression-gaps}
Despite the significant achievements of Transformer models in various NLP tasks, research in the area of time expression identification and normalization has remained relatively limited over the past few years. This is particularly true of time normalization, where the volume and depth of research are low, especially when compared to similar tasks such as named entity recognition, entity normalization, and entity linking. Furthermore, the methodological diversity in existing works is notably constrained, with most research relying on pre-trained Transformer models for simple token classification. While generative LLMs like GPT-4 or LLAMA3 have demonstrated impressive performance in other NLP tasks, their potential in the identification and normalization of time expressions has barely been explored. This suggests a significant research gap exists; exploration of generative approaches may offer the potential for advancement in time expression identification and normalization.


\section{Temporal Relation Methods}
\label{sec:temporal-relation-extraction}
The task of temporal relation extraction typically assumes that events and time expressions in the text have already been identified, with the only objective being to extract the temporal relations between them. 
We summarize all the reviewed temporal relation extraction works in \Cref{sec:temporal-relation-extraction-methods-summary} \Cref{table:temporal-relation-extraction}.
Discriminative methods typically employ a pretrained discriminative language model like BERT or RoBERTa \cite{liu2019roberta} as the base encoder model to derive contextualized representations of events or time expressions. Subsequently, these representations are paired and input into a classification layer for a multi-class classification task, with each class representing a different temporal relation. Generative methods typically leverage encoder-decoder models such as T5 \cite{raffel2020exploring} or decoder-only models like GPT \cite{radford2019language} to generate a target sequence that encapsulates the temporal relation between the input events and times. These methods often rely on post-processing techniques to extract specific temporal relations from the predicted target sequences.

\subsection{Discriminative Methods Overview}

Works on discriminative temporal relation extraction have mainly focused on integrating external knowledge and improving model robustness.

\subsubsection{Integrating External Knowledge}

\paragraph{Commonsense Knowledge} 
Commonsense knowledge for temporal relations usually involves typical sequences of events, such as eating typically occurring after cooking. Such commonsense knowledge might be fundamental for humans, but absent from the base encoder model. 
\citet{ning-etal-2019-improved}, \citet{wang-etal-2020-joint} and \citet{tan-etal-2023-event} integrated knowledge from external commonsense knowledge graphs. \citet{tan-etal-2023-event} employs a complex Bayesian learning method to merge the knowledge with the contextualized representations from the base encoder, whereas \citet{ning-etal-2019-improved} and \citet{wang-etal-2020-joint} simply concatenate the vectorized representations of the commonsense knowledge with those from the base encoder.

\paragraph{Syntactic and Semantic Knowledge}
Syntactic and semantic knowledge, typically extracted using off-the-shelf external tools or straightforward rules, enrich the base encoder models' representations. For instance, \citet{wang-etal-2022-dct} utilizes SpaCy's dependency parser to parse the syntactic dependency trees from the input text and neuralcoref to identify coreferential relationships among entities. \citet{mathur-etal-2021-timers} employs the discoursegraphs library to parse rhetorical dependency graphs from the text. To integrate this structured knowledge into the contextualized event or time expression representations, graph neural networks are often employed over syntactic or semantic pairwise relations \cite{wang-etal-2022-dct, mathur-etal-2022-doctime, zhou-etal-2022-rsgt,mathur-etal-2021-timers}. For example, \citet{wang-etal-2022-dct} first encodes an input sequence containing event pairs with the RoBERTa model to generate initial contextual representations, which are then enhanced with extracted syntactic and semantic knowledge using additional graph neural network layers. Another method is to prelearn or extract vectorized representations of the knowledge, which are later concatenated with the event or time expression representations \cite{ross-etal-2020-exploring,wang-etal-2020-joint,han-etal-2019-deep,ning-etal-2019-improved,han2019contextualized}, as in \citet{wang-etal-2020-joint}, where RoBERTa token embeddings and one-hot vectors of part-of-speech tags are combined.

\paragraph{Temporal-Specific Rules}
These rules are intrinsic to temporal relations themselves, with symmetry and transitivity being the most common.
For instance, if event A happens before event B, then symmetry can be used to infer that B happens after A.
And if A precedes B and B precedes C, transitivity can be used to infer that A precedes C. 
Detailed explanations of the symmetry and transitivity rules and a comprehensive transitivity table are provided in \citet{ning-etal-2019-improved}.
Such rules can be embedded during the model training phase, enabling the model to learn the characteristics of these temporal relations. \citet{hwang-etal-2022-event} and \citet{tan-etal-2021-extracting} utilize box embedding and hyperbolic embedding, respectively, to implicitly guide the model in understanding and learning the symmetry and transitivity rules. \citet{zhou2021clinical} and \citet{wang-etal-2020-joint} translate the constraints of temporal relations into regularization terms for the loss function during training to penalize predictions that violate these rules.
Alternatively, rules can be embedded during the inference phase to ensure that all deduced temporal relations adhere to the symmetry and transitivity rules as closely as possible. Custom heuristics in \citet{wang-etal-2022-dct,zhou-etal-2022-rsgt,zhou2021clinical,liu2021discourse} exclude temporal relations that contravene rules during inference. \citet{wang-etal-2020-joint} and \citet{han-etal-2019-joint} formulate the inference of temporal relations as a linear programming problem, optimizing the solution to achieve optimal outcomes. \citet{han-etal-2019-deep} interprets the discriminative model's output probabilities as confidence scores for potential relations between temporal entity pairs and employs a structured support vector machine for the final predictions.

\paragraph{Label Distribution}
Knowledge of label distribution pertains to the frequency distribution of specific temporal relations in the training set. \citet{wang-etal-2023-extracting} and \citet{han-etal-2020-domain} integrate this distribution knowledge into their frameworks, using it as a regularization term in the loss function or for inference-time linear programming, aiming to mitigate potential biases in model predictions.

\subsubsection{Improving Model Robustness}

\paragraph{Multitask Learning}
\citet{wang-etal-2022-dct}, \citet{lin-etal-2020-bert} and \citet{cheng-etal-2020-dynamically} categorize temporal relations and treat the extraction of different types of temporal relations as independent tasks, employing multitask learning to extract all types of relations simultaneously. For instance, \citet{wang-etal-2022-dct} delineates tasks into event-event, event-time, and event-document creation time, undergoing multitask training across these three tasks. 
\citet{mathur-etal-2022-doctime} applies multitask learning in their model to concurrently predict temporal relations and dependency links between nodes in a temporal dependency tree. 
Similarly, \citet{ballesteros-etal-2020-severing} implements multitask learning by integrating the extraction of temporal relations with the extraction of entity relations in the general domain.

\paragraph{Data Augmentation}
\citet{wang-etal-2023-extracting} generates counterfactual instances from the training set samples to mitigate model bias, while \citet{tiesen-lishuang-2022-improving} employs predefined templates to create additional training examples.

\paragraph{Continued Pre-training of Base Encoder}
In \citet{zhao-etal-2021-effective} and \citet{han-etal-2021-econet}, heuristic methods are used to identify temporal indicators in a corpus of unlabeled data, further training the base encoder using a masked language modeling (MLM) approach to recover masked indicators. \citet{lin-etal-2019-bert} focuses on the medical domain, using MLM on electronic health records from MIMIC-III to adapt the base encoder for domain-specific training prior to temporal relation extraction.

\paragraph{Adversarial Training}
\citet{kanashiro-pereira-2022-attention} and \citet{pereira-etal-2021-alice} introduce adversarial perturbations at different layers of the transformer encoder during training to enhance model robustness.

\paragraph{Self-training}
\citet{cao2021uncertainty} and \citet{ballesteros-etal-2020-severing} initially train a temporal relation extraction model on annotated datasets and then apply the model to unlabeled data to obtain model-generated labels as pseudo labels. They subsequently select pseudo-labeled examples as sliver examples based on the model's uncertainty scores and confidence scores (probability scores for specific temporal relation predictions) to train the model.


\subsection{Generative Methods Overview}
Unlike the task of extracting relations between general entities for constructing knowledge graphs (refer to survey \citet{ye-etal-2022-generative}), few generative approaches have been proposed and applied in the field of temporal relation extraction.
\citet{dligach-etal-2022-exploring} utilizes an encoder-decoder model architecture, specifically the BART \cite{lewis-etal-2020-bart} and T5 \cite{raffel2020exploring} models. They primarily investigate how to fine-tune these encoder-decoder models for temporal relation extraction tasks, focusing on the input and output formats. They discover that producing outputs for each event and time pair separately is more effective than the intuitive triplet form, i.e., (entity, relation, entity). On the other hand, \citet{yuan-etal-2023-zero} concentrates on examining the capabilities of the powerful ChatGPT generative model, in the context of temporal relation extraction, testing various prompting methods, such as zero-shot prompting, and the popular chain-of-thought prompting \cite{wei2022chain}. Their findings indicate that, despite using these prompting methods, ChatGPT's performance in temporal relation extraction still falls significantly short compared to fine-tuned transformer-based models.

\subsection{Discussion and Research Gaps}
\label{sec:temporal-relation-gaps}
\paragraph{Homogenization of Methods and Evaluations}
While numerous Transformer-based methods for temporal relation extraction have emerged, they tend to be algorithmically similar, utilizing discriminative base models like BERT to represent temporal entities and incorporating additional knowledge into these representations. A common strategy involves using off-the-shelf IE tools to extract syntactic knowledge and enhance the base model's representations with graph neural networks. The small gains in state-of-the-art performance from one model to the next probably represent additional hyperparameter tuning more than substantial progress in understanding the relations between temporal entities in text.

Most works also focus on only three datasets -- MATRES, TimeBank-Dense, and TDDiscourse -- which are predominantly in the newswire domain with only 274, 36, and 34 documents, respectively, and exhibit significant overlap. This limitation in datasets might lead to an incomplete assessment of the models' generalization capabilities. Repeated testing and fine-tuning on these small, overlapping datasets could result in overfitting, failing to reflect the models' effectiveness on broader and more diverse datasets. Moreover, this singular domain-focused evaluation approach could cause severe domain bias, leaving the applicability of these methods outside the news domain uncertain.

\paragraph{Absence of Generative LLMs}
In temporal relation extraction, we observe a phenomenon similar to that in time expressions---there is a lack of applications using generative LLMs, which have shown excellent performance in natural language processing tasks. While there are two works that attempt to explore Transformer-based generative approaches, they are limited to studying different formats in input and output. We have not seen further exploration or application of more complex prompting techniques or training strategies.

\paragraph{Increased Demand for Model Openness}
As shown in the last column of \Cref{table:temporal-relation-extraction}, most temporal relation extraction models are not publicly available, possibly due to the absence of code releases or the need to re-train models on new datasets even when code is provided. Re-training a model involves significant replication work.
This inaccessibility directly impacts the practical application and testing of these trained models in other temporal reasoning tasks, thereby affecting the development of the temporal relation extraction field. Given the application-oriented nature of temporal relation extraction tasks, only by understanding the specific issues encountered in actual applications can we propose strategies to address these real-world challenges.

\section{Applications}
\label{sec:Temporal-Information-Extraction-Application}
\subsection{Methods Overview}
Temporal IE is often regarded as an ``upstream'' system, akin to other general IE systems. These systems aim to extract structured information to improve the reasoning of ``downstream'' tasks, such as temporal reasoning. A natural question is how the models from  \Cref{sec:time-expression,sec:temporal-relation-extraction} are used in downstream tasks to help temporal reasoning.

Despite a wealth of research on Transformer-based temporal IE systems in recent years, there has been scant application of these systems' outputs in temporal reasoning tasks. Only a few temporal reasoning tasks, such as timeline summarization and temporal question answering, leverage the results of temporal IE. 
The timeline summarization task aims to chronologically order and label key dates of events within a collection of news documents, while temporal question answering relies on unstructured context documents to answer temporal-related questions. Both tasks require reasoning about time and events to generate outcomes.

One approach to utilizing temporal IE systems is to explicitly construct temporal graphs to assist with temporal reasoning.
Some works use only simple temporal graphs containing only time expressions extracted by rules \cite{su-etal-2023-fusing} or transformers \cite{yang-etal-2023-upon,xiong2024large} and normalized by rules.
Other works use complete temporal graphs constructed by a complete temporal IE pipeline, including time expression identification, normalization, and temporal relation extraction, with \citet{mathur-etal-2022-doctime} using Transformer-based relation extraction, and \citet{li-etal-2021-timeline} using LSTM-based relation extraction and rules for the other components.
As for the usage of the constructed temporal graph, they can be input into models directly in text form \cite{su-etal-2023-fusing,yang-etal-2023-upon,xiong2024large} or encoded into the hidden states of a Transformer model through an attention fusion mechanism or graph neural networks \cite{li-etal-2021-timeline,mathur-etal-2022-doctime,su-etal-2023-fusing}.

Some works only preprocess the input with a specific temporal IE component rather than building a temporal graph. For instance, \citet{bedi-etal-2021-temporal} employs the rule-based HeidelTime \cite{strotgen-gertz-2010-heideltime} for extracting and normalizing time expressions in texts for constructing the input of a temporal question generation model; while \citet{cole-etal-2023-salient} uses the rule-based SUTime \cite{chang-manning-2012-sutime} to process the entire Wikipedia, supporting the temporal pre-training of the Transformer model.

\subsection{Discussion and Research Gaps}
\label{sec:applications-gaps}
Although there is considerable work on transformer-based temporal IE, especially in temporal relation extraction tasks, these methods have not been widely applied to downstream tasks. 
For example, there are many Transformer-based works that have been trained on the MATRES dataset, but none have been utilized in downstream tasks. This may be attributed to most temporal IE models not being publicly available, as shown in \Cref{table:temporal-relation-extraction}. Replicating these models can be both complex and time-consuming, requiring substantial effort.
Furthermore, existing models exhibit domain bias. For example, in temporal relation extraction tasks, most research relies on the TimeBank-Dense and MATRES datasets, which primarily contain data from the newswire domain. Hence, the generalization capabilities of these models in other domains might be limited.

\section{Conclusion}
In this paper, we provide an overview of three classic tasks in the field of temporal IE: time expression identification, time expression normalization, and temporal relation extraction. We discuss datasets, Transformer-based methods, and their applications within these areas. We found that although Transformer models have demonstrated outstanding performance on many NLP tasks, there remain significant research gaps in the domain of temporal IE. For example, there is a noticeable lack of studies involving LLMs. We hope this survey will offer a comprehensive review and insights to researchers in the field, inspiring further research to address these existing gaps. We expand on the research opportunities arising from these gaps in \Cref{sec:discussion-on-future-directions}.

\section*{Limitations}
\label{sec:limitations}
In this review, we focus exclusively on transformer-based temporal IE methods, without including rule-based approaches. We also center our discussion on the most common temporal IE tasks rather than addressing every possible subtask.

\bibliography{anthology,custom}
\bibliographystyle{acl_natbib}

\appendix

\section{Evaluation Metrics}
\label{sec:eval-metrics}
In temporal IE, the evaluation method from TEMPEVAL-3 \cite{uzzaman-etal-2013-semeval} is the most widely adopted standard. This evaluation method calculates the standard precision (P), recall (R), and F1 score (F) between the system predictions (System) and the gold annotations (Reference) as follows:

\begin{equation}
    P = \frac{|\text{System} \cap \text{Reference}|}{|\text{System}|}
\end{equation}
\begin{equation}
    R = \frac{|\text{System} \cap \text{Reference}|}{|\text{Reference}|}
\end{equation}
\begin{equation}
    F = 2 \cdot \frac{P \cdot R}{P + R}
\end{equation}

In time expression identification, ``System'' refers to the time expressions identified by the system, while ``Reference'' refers to the annotated gold time expressions. 
In time expression normalization, ``System'' and ``Reference'' refer to the system-normalized time expressions and the gold annotated normalized expressions, respectively. If calculating the end-to-end time expression normalization score, ``System'' only involves the correctly identified time expressions.

For the temporal relation extraction task, the TEMPEVAL-3 evaluation method calculates the temporal awareness scores. This is achieved by performing a graph closure operation on the gold temporal graph based on temporal transitivity rules (to incorporate all potential temporal relations) and reducing the predicted temporal relation graph (to remove duplicate relations). These steps are completed before calculating the standard scores. Here, ``System'' denotes the temporal relations predicted by the system, while ``Reference'' is the gold annotated temporal relations.

\section{Datasets Summary}
\label{sec:datasets-summary}
We summarize the temporal IE datasets in \Cref{table:dataset-overview}. The first section is based on the most widely used TimeML annotation framework, while the second section covers those that adopt all other annotation frameworks.

\begin{table*}[t!]
\centering
\small
\setlength{\tabcolsep}{.5em}
\begin{tabular}{@{} l p{2.3cm} p{2.6cm} p{1.6cm} l @{}}
\toprule
\textbf{Name} & \textbf{Framework} & \textbf{Domain} & \textbf{Lang} & \textbf{Tasks} \\
\midrule
\multicolumn{5}{c}{\emph{TimeML-Based}} \\
\midrule
TimeBank \cite{james2003timeml} & TimeML & Newswire & EN & I, N, R \\
TempEval-1 \cite{verhagen-etal-2007-semeval} & TimeML & Newswire & EN & I, N, R \\
TempEval-2 \cite{verhagen-etal-2010-semeval} & TimeML & Newswire & ZH, EN, IT, FR, KR, ES & I, N, R \\
Spanish TimeBank \cite{nieto2011modes} & TimeML & Historiography & ES & I, N \\
French TimeBank \cite{bittar-etal-2011-french} & ISO-TimeML & Newswire & FR & I, N, R \\
Portuguese TimeBank \cite{costa-branco-2012-timebankpt} & TimeML & Newswire & PT & I, N, R \\
i2b2-2012 \cite{sun2013evaluating} & Thyme-TimeML & Clinical & EN & I, N, R \\
TempEval-3 \cite{uzzaman-etal-2013-semeval} & TimeML & Newswire & EN, ES & I, N, R \\
TimeBank-Dense \cite{chambers-etal-2014-dense} & TimeML & Newswire & EN & I, N, R \\
Japanese TimeBank \cite{asahara-etal-2013-bccwj} & ISO-TimeML & Publication, Library, Special purpose & JA & I, N, R \\
AncientTimes \cite{strotgen-etal-2014-extending} & TimeML & Wikipedia & EN, DE, NL, ES, FR, IT, AR, VI & I, N \\
THYME-2015 \cite{bethard-etal-2015-semeval} & Thyme-TimeML & Clinical & EN & I, N, R \\
THYME-2016 \cite{bethard-etal-2016-semeval} & Thyme-TimeML & Clinical & EN & I, N, R \\
Richer Event Description \cite{ogorman-etal-2016-richer} & Thyme-TimeML & Newswire, Forum Discussions & EN & I, N, R \\
Italian TimeBank \cite{bracchi2016enrichring} & TimeML & Newswire & IT & I, N, R \\
MeanTime \cite{minard-etal-2016-meantime} & ISO-TimeML & Newswire & EN, IT, ES, NL & I, N, R \\
THYME-2017 \cite{bethard-etal-2017-semeval} & Thyme-TimeML & Clinical & EN & I, N, R \\
Event StoryLine \cite{caselli-vossen-2017-event} & TimeML & Story & EN & I, N, R \\
MATRES \cite{ning-etal-2018-multi} & TimeML & Newswire & EN & I, R \\
Korean TimeBank \cite{lim-etal-2018-korean} & TimeML & Wikipedia & KR & I, N, R \\
German Temporal Expression \cite{strotgen-etal-2018-krauts} & TimeML & Newswire & DE & I, N \\
TDDiscourse \cite{naik-etal-2019-tddiscourse} & TimeML & Newswire & EN & R \\
PATE \cite{zarcone-etal-2020-pate} & TimeML & Voice Assistant & EN & I, N \\
German VTEs \cite{may-etal-2021-extraction} & ISO-TimeML & Newswire & DE & I, N \\
\midrule
\multicolumn{5}{c}{\emph{Other Annotation Framework-based}} \\
\midrule
WikiWars \cite{mazur-dale-2010-wikiwars} & TIMEX2 & Wikipedia & EN, DE & I, N \\
SCATE \cite{bethard-parker-2016-semantically,laparra-etal-2018-semeval} & SCATE & Newswire, Clinical & EN & I, N \\
CaTeRS \cite{mostafazadeh-etal-2016-caters} & CaTeRS & Commonsense Stories & EN & R \\
TORDER \cite{cheng-miyao-2018-inducing} & TORDER & Newswire & EN & R \\
Temporal Dependency Tree \cite{zhang-xue-2018-structured,zhang-xue-2019-acquiring} & Temporal Dependency Tree & Newswire, Narratives & ZH & R \\
Temporal Dependency Graph \cite{yao-etal-2020-annotating} & Temporal Dependency Graph & Newswire & EN & R \\
\bottomrule
\end{tabular}
\vspace{-.5\baselineskip}
\caption{Overview of datasets and their schemas, domains, languages (EN: English, DE: German, NL: Dutch, ES: Spanish, FR: French, IT: Italian, AR: Arabic, VI: Vietnamese, JA: Japanese, PT: Portuguese, ZH: Chinese, KR: Korean), and tasks (I: identification, N: time expression normalization, R: temporal relation extraction).}
\label{table:dataset-overview}
\end{table*}

\section{Temporal Relation Extraction Methods Summary}
\label{sec:temporal-relation-extraction-methods-summary}
We summarize the temporal relation extraction methods we review in \Cref{table:temporal-relation-extraction}.

\begin{table*}[t!]
\small
\centering
\setlength{\tabcolsep}{.5em}
\begin{tabular}{l l l p{4cm} l l l}
\toprule
\textbf{Work} & \textbf{Approach} & \textbf{Base Model} & \textbf{Evaluation Datasets} & \textbf{Knowl.} & \textbf{Robust} & \textbf{Avail.} \\
\midrule
\citet{lin-etal-2019-bert} & Discr. & BERT & THYME & \XSolidBrush & \Checkmark & \XSolidBrush \\
\citet{han-etal-2019-deep} & Discr. & BERT & TimeBank-Dense, MATRES & \Checkmark & \XSolidBrush & \XSolidBrush \\
\citet{ning-etal-2019-improved} & Discr. & BERT & TimeBank-Dense, MATRES & \Checkmark & \XSolidBrush & \XSolidBrush \\
\citet{han-etal-2019-joint} & Discr. & BERT & TimeBank-Dense, MATRES & \Checkmark & \Checkmark & \XSolidBrush \\
\citet{han2019contextualized} & Discr. & BERT & Richer Event Description, CaTeRS & \Checkmark & \Checkmark & \XSolidBrush \\
\citet{lin-etal-2020-bert} & Discr. & BERT & THYME & \XSolidBrush & \Checkmark & \XSolidBrush \\
\citet{cheng-etal-2020-dynamically} & Discr. & BERT & Japanese-Timebank, TimeBank-Dense & \Checkmark & \Checkmark & \XSolidBrush \\
\citet{ross-etal-2020-exploring} & Discr. & BERT & Temporal Dependency Tree & \Checkmark & \XSolidBrush & \XSolidBrush \\
\citet{ballesteros-etal-2020-severing} & Discr. & RoBERTa & MATRES & \XSolidBrush & \Checkmark & \XSolidBrush \\
\citet{han-etal-2020-domain} & Discr. & RoBERTa & i2b2-2012, TimeBank-Dense & \Checkmark & \Checkmark & \XSolidBrush \\
\citet{wang-etal-2020-joint} & Discr. & RoBERTa & MATRES & \Checkmark & \XSolidBrush & \XSolidBrush \\
\citet{zhao-etal-2021-effective} & Discr. & RoBERTa & MATRES & \XSolidBrush & \Checkmark & \Checkmark \\
\citet{zhou2021clinical} & Discr. & BERT & i2b2-2012, TimeBank-Dense & \Checkmark & \XSolidBrush & \XSolidBrush \\
\citet{cao2021uncertainty} & Discr. & RoBERTa & MATRES, TimeBank-Dense & \XSolidBrush & \Checkmark & \XSolidBrush \\
\citet{tan-etal-2021-extracting} & Discr. & RoBERTa & MATRES & \Checkmark & \XSolidBrush & \XSolidBrush \\
\citet{mathur-etal-2021-timers} & Discr. & BERT & TimeBank-Dense, MATRES, TDDiscourse & \Checkmark & \XSolidBrush & \XSolidBrush \\
\citet{liu2021discourse} & Discr. & BERT & TimeBank-Dense, TDDiscourse & \Checkmark & \XSolidBrush & \XSolidBrush \\
\citet{wen-ji-2021-utilizing} & Discr. & RoBERTa & MATRES & \Checkmark & \XSolidBrush & \XSolidBrush \\
\citet{pereira-etal-2021-alice} & Discr. & RoBERTa & MATRES, TimeML & \XSolidBrush & \Checkmark & \XSolidBrush \\
\citet{han-etal-2021-econet} & Discr. & RoBERTa/BERT & TimeBank-Dense, MATRES, Richer Event Description & \XSolidBrush & \Checkmark & \Checkmark \\
\citet{kanashiro-pereira-2022-attention} & Discr. & RoBERTa & MATRES, TimeML & \XSolidBrush & \Checkmark & \XSolidBrush \\
\citet{wang-etal-2022-dct} & Discr. & RoBERTa & TimeBank-Dense, TDDiscourse & \Checkmark & \Checkmark & \XSolidBrush \\
\citet{mathur-etal-2022-doctime} & Discr. & BERT & Temporal Dependency Tree & \Checkmark & \Checkmark & \XSolidBrush \\
\citet{hwang-etal-2022-event} & Discr. & RoBERTa & MATRES, Event StoryLine & \Checkmark & \XSolidBrush & \XSolidBrush \\
\citet{dligach-etal-2022-exploring} & Gen & BART/T5 & THYME & \XSolidBrush & \XSolidBrush & \XSolidBrush \\
\citet{wang-etal-2023-extracting} & Discr. & BigBird & MATRES, TDDiscourse & \Checkmark & \Checkmark & \XSolidBrush \\
\citet{zhang-etal-2022-extracting} & Discr. & BERT & MATRES, TimeBank-Dense & \Checkmark & \XSolidBrush & \XSolidBrush \\
\citet{tiesen-lishuang-2022-improving} & Discr. & BERT & TimeBank-Dense, MATRES & \XSolidBrush & \Checkmark & \XSolidBrush \\
\citet{zhou-etal-2022-rsgt} & Discr. & RoBERTa & TimeBank-Dense, MATRES & \Checkmark & \XSolidBrush & \XSolidBrush \\
\citet{man2022selecting} & Discr. & RoBERTa & MATRES, TDDiscourse & \Checkmark & \XSolidBrush & \XSolidBrush \\
\citet{yuan-etal-2023-zero} & Gen & ChatGPT & TimeBank-Dense, MATRES, TDDiscourse & \XSolidBrush & \XSolidBrush & \XSolidBrush \\
\citet{tan-etal-2023-event} & Discr. & BART & MATRES, imeBank-Dense & \Checkmark & \XSolidBrush & \Checkmark \\
\bottomrule
\end{tabular}
\caption{Overview of research on temporal relation extraction. ``Knowl.'' represents the inclusion of external knowledge. ``Robust" refers to the application of methods to enhance model robustness. ``Avail.'' indicates whether the model is publicly available. Symbols \Checkmark and \XSolidBrush indicate the presence or absence of a feature, respectively.}
\label{table:temporal-relation-extraction}
\end{table*}

\section{Discussion on Future Directions}
\label{sec:discussion-on-future-directions}

In the previous sections, we have identified the following research opportunities in the field of temporal IE:
\begin{itemize}
    \item Enrich annotation frameworks (\Cref{sec:datasets-gaps}), e.g., representing event arguments or expanding formal semantic systems like SCATE.
    \item Improve dataset diversity (\Cref{sec:datasets-gaps}), e.g., annotating more domains beyond newswire.
    \item Explore generative approaches (\Cref{sec:time-expression-gaps,sec:temporal-relation-gaps}), e.g., new input-output formulations, new fine-tuning strategies.
    \item Develop public tools and benchmarks (\Cref{sec:time-expression-gaps,sec:temporal-relation-gaps}), e.g., publish temporal IE models and datasets to the public repositories
    \item Explore new applications (\Cref{sec:applications-gaps}), e.g., the utility of extracted timelines when visualized for human-computer interaction.
\end{itemize}

\subsection{Enrich Annotation Frameworks and Improve the Domain Diversity of Datasets}
Current annotation frameworks, such as TimeML, often produce temporal graphs composed of temporal relations and temporal entities, as illustrated in \Cref{fig:temporal-graph}. However, these temporal graphs are challenging to interpret independently or use directly for temporal reasoning without extensive context. One future direction could be to integrate richer content into end-to-end temporal IE annotation frameworks. One example is incorporating entity relation extraction and full event extraction (including triggers and arguments) from the general domain to construct a more complete temporal graph. This concept has begun to emerge in the literature, as seen in \citet{li-etal-2021-timeline}. Yet, that work mainly integrates existing temporal IE tools with general domain IE tools without proposing a well-defined annotation framework. Another example is to develop user-friendly frameworks like SCATE, which, unlike TimeML, outputs temporal intervals that can be directly mapped onto a timeline given a temporal expression. However, SCATE primarily focuses on the normalization of time expressions. Expanding its scope to include the normalization of a broader range of temporal content, such as events and sentences, could significantly widen its applicability.

Furthermore, future efforts could focus on expanding the domains covered by existing datasets to mitigate the domain bias present in current datasets. For example, the Thyme datasets represent an adaptation of TimeML to better suit the medical field's representation of temporal relations between events and times. Yet, such efforts to adapt and improve annotation frameworks for additional fields are still scarce. Therefore, adapting existing annotation frameworks to a broader range of domains to enhance the domain diversity of datasets represents a potential future research direction.

\subsection{Improve the Application of Generative LLMs}

The application of generative LLMs in the field of time expression identification, normalization, and temporal relation extraction remains underexplored. 
Given the proven capabilities of LLMs like ChatGPT and LLAMA3 across various tasks, it is logical to probe their potential within the realm of temporal IE. Whether it involves leveraging new prompting methods or fine-tuning strategies for specific tasks, there is ample room for innovation.

However, it is important to emphasize that while these models excel in generating unstructured text when applied to temporal IE, it is imperative to specially design suitable input-output formats. Such designs are intended to enable generative LLMs, which are typically used for producing unstructured text, to also effectively output structured temporal information.

\subsection{Develop Public Toolkits and Evaluation Benchmarks}
We believe that one key reason transformer-based temporal IE models have not been widely adopted might be the absence of a publicly available code repository that facilitates easier access to models and data. 
For example, HuggingFace \footnote{\url{https://huggingface.co/}} provides language model heads or pipelines suitable for various tasks, allowing users to easily download and deploy trained models on any dataset directly from the HuggingFace Hub. A future research direction should involve establishing such a repository or pushing models/datasets to HuggingFace Hub for the temporal IE tasks to enhance the reproducibility and applicability of research. Another important direction is to create a public and test-set concealed benchmark for a more equitable comparison of existing work. In most existing works, although metrics such as F1 scores, precision, and recall are commonly computed, the specific implementations can vary. For instance, in \citet{kanashiro-pereira-2022-attention}, only the ``before'' and ``after'' relationships are evaluated for relation extraction performance, whereas \citet{zhang-etal-2022-extracting} includes all temporal relationships except ``vague'' in their evaluation.

\subsection{Explore More Application Directions}
\label{subsec:hci}
In reviewing the application of temporal IE systems, we observe that current research primarily focuses on aiding ``models'' in temporal reasoning to enhance their performance in other tasks. Future research in temporal IE should not only continue to support model performance improvement but should also pay more attention to serving humans and enhancing its practical value. A promising application direction is visualizing timelines in human-computer interaction (HCI) scenarios. The visualization results of existing temporal graphs are often challenging for human users to interpret. For instance, visualizing the temporal graph of any document in the TimeBank-Dense dataset might result in a graph densely populated with points and lines, offering little help for users to comprehend the progression of events within the text.

User studies, such as those conducted by \citet{di2020evaluating}, have revealed the importance of visualization forms of timelines for user understanding. Consequently, temporal IE research should also consider incorporating user research on temporal graphs to guide the design of temporal IE methods, such as how to represent standardized time expressions, identify which types of temporal relations most effectively facilitate time understanding, and determine the best ways to present this information. By addressing these problems, the extraction and representation of temporal information can be more closely aligned with user needs, enhancing its application value in HCI.

\end{document}